\documentclass[journal]{IEEEtran}
\usepackage[T1]{fontenc}
\usepackage[utf8]{inputenc}

\usepackage{amsmath,amssymb,amsfonts,amsthm,bm} 
\usepackage{algorithmic}
\usepackage{algorithm}
\usepackage{array,multirow,makecell}
\usepackage{textcomp}
\usepackage{stfloats}
\usepackage{url}
\usepackage{verbatim}
\usepackage{graphicx}
\usepackage{cite}
\usepackage{tabularx,booktabs,colortbl}
\usepackage{caption,subcaption}
\usepackage{adjustbox}
\usepackage[switch]{lineno}
\usepackage[hidelinks]{hyperref}
\usepackage[per-mode=symbol]{siunitx}
\usepackage{tikz}
\usepackage{dsfont}

\definecolor{highlightColor}{RGB}{112,140,156} 
\graphicspath{{images/}}
\newcounter{assno}

\newcommand{\numberset}[1]{\mathds{#1}}
\newcommand{\R}{\numberset{R}}
\begin{document}

\title{A Human-in-the-Loop Comparison of Proportional Control and Adaptive Gravity Compensation in Elbow EMG-Driven Exoskeletons}

\author{Francesco Pascucci$^{1}$, Eldison Dimo$^{1}$, Davide Costanzi$^{1}$, Mario Sedilesu$^{1}$, Andrea Calanca$^{1}$

\thanks{$^{1}$Department of Engineering for Innovation Medicine, University of Verona, Italy.
Strada Le Grazie 15, 37134 Verona, Italy. E-mails: \texttt{\small{} francesco.pascucci\_01@univr.it}, \texttt{\small{} eldison.dimo@univr.it}, \texttt{\small{} mario.sedilesu@univr.it}, \texttt{\small{} davide.costanzi@univr.it}, \texttt{\small{} andrea.calanca@univr.it}. The corresponding author is Francesco Pascucci.}
}

\maketitle

\begin{abstract}
Electromyography (EMG)-driven control enables wearable robots to assist users based on their neuromuscular activity. Upper-limb exoskeletons typically rely on proportional myoelectric control (PMC), which amplifies user effort but may compromise stability at high assistance levels. EMG-driven adaptive gravity compensation (AGC) offers a promising alternative, counterbalancing gravitational torques while continuously adapting to neuromuscular activity. However, these approaches have not been systematically compared in a human-in-the-loop setting. This study evaluates five control algorithms: two PMC strategies at different assistance levels, a purely integral AGC (\textit{AGC-PI}), a kinematics- and acceleration-aware model-based (MB) AGC (\textit{AGC-AW}), and a novel MB AGC approach based on Recursive Least Squares (\textit{AGC-RLS}) for online payload estimation. Fourteen healthy participants performed target-reaching tasks using an active elbow exoskeleton under 0~\unit{kg} and 2~\unit{kg} loading conditions. Performance metrics included muscular activity, movement accuracy, task efficiency, and subjective workload. PMC exhibited a trade-off between physical effort and task performance: low assistance levels required greater physical effort under load, whereas high gains impaired task execution. \textit{AGC-PI} reduced agonist activity but induced antagonist activation. In contrast, both model-based AGC approaches effectively compensated for external payloads, reducing physical and cognitive demands while improving movement accuracy and completion time.
\end{abstract}




\section{Introduction}\label{sec:intro}
\IEEEPARstart{U}{pper-limb} active exoskeletons have the potential to substantially reduce physical effort, enhance task performance, and support motor function across a broad spectrum of applications, spanning industrial settings, activities of daily living, and clinical rehabilitation \cite{Gull2020,Pascucci2025}. Among the available control paradigms, electromyography (EMG)-driven strategies have attracted considerable attention because they establish a direct interface with the user's neuromuscular activity, enabling intuitive and responsive assistance \cite{Fu2022ReviewMCS}. By exploiting EMG signals, these approaches can infer the user’s movement intent before any observable motion occurs, allowing the exoskeleton to proactively support the task \cite{robert1979:emd}. Among the various EMG-driven control paradigms proposed in the literature, this work focuses on two: proportional myoelectric control (PMC), the most widely adopted strategy in assistive exoskeletons, and EMG-driven adaptive gravity compensation (AGC), an emerging class of assistive controllers.

In the PMC paradigm, the assistive torque is proportional to the muscle activation of the user, typically through a gain-based mapping (that defines the level of assistance) between the processed EMG signals and the actuator torque \cite{lenzi2012:emg,lu_myoelectric_2019,lotti2022:myo_vs_force,Peng2024Myomo}. These approaches provide intuitive assistance that directly amplifies the user’s effort, which explains their widespread adoption in the literature. However, their performance strongly depends on gain tuning, and too high gains may degrade usability and system stability \cite{Kwon2014}.

Given that most of the effort in upper-limb motions is spent against gravity, AGC strategies assist the user by generating torque that compensates for the limb's gravitational effects. Some implementations have been proposed that exploit EMG signals to implicitly or explicitly estimate such gravitational forces \cite{TreussartVignais2020,Nasiri2022,Dimo2026MyographyGravityCompensation,Sierotowicz2022:paexo_active}. AGC approaches have only recently emerged, and they offer a conceptually distinct mechanism of assistance that makes them inherently capable of providing near-complete assistance, a regime that proportional controllers cannot sustain without compromising stability.

Despite the fundamental differences between these two paradigms, no systematic comparison of their relative merits exists. Broader comparative studies in the EMG-driven exoskeleton literature have addressed related but distinct questions: the contrast between EMG-driven and model-based control \cite{lotti2020:myocontrol_soft}, the effect of assistance magnitude on muscle activity in PMC \cite{missiroli2020:assistance_level}, and the comparison of PMC with biological torque-based strategies in a hip exoskeleton \cite{young2017:hip_exo}. The specific question of which EMG-driven paradigm better serves the user and under what conditions remains open.

This paper addresses this gap by presenting a human-in-the-loop experimental comparison of PMC and AGC strategies implemented on an active elbow exoskeleton. The specific contributions of this work are twofold. First, we propose a novel model-based (MB) AGC strategy based on Recursive Least Squares (\textit{AGC-RLS}) for online adaptation of the gravity-compensating assistive torque. Second, we systematically evaluate this approach against two PMC and two state-of-the-art AGC algorithms using a multidimensional assessment with 14 healthy participants under two loading conditions.



\section{Background}\label{sec:background}

\subsection{Human–Exoskeleton-Payload Dynamic Model}

The human-exoskeleton-payload system is modeled as a single-degree-of-freedom (1-DoF) mechanism for elbow flexion–extension:
\begin{equation}\label{eq:system}
    (m_h b_h^2 + m_p b_p^2)\ddot{\theta} 
    + (m_h b_h + m_p b_p) g \cos\theta 
    = \tau_h + \tau_e .
\end{equation}
Here, $\theta$ denotes the absolute orientation of the forearm with respect to the ground, with $\theta=0$ when the forearm is parallel to the ground. The terms $m_h$ and $m_p$ denote the mass of the forearm–exoskeleton system and the handled payload, while $b_h$ and $b_p$ are their respective center-of-mass (COM) distances from the joint axis. $\tau_h$ and $\tau_e$ represent the torques generated by the wearer’s muscles and by the exoskeleton.


\subsection{EMG-Driven Exoskeleton Control}\label{subsec:ExoContro}

\begin{figure}[tp]
    \centering
    \includegraphics[width=0.9\columnwidth]{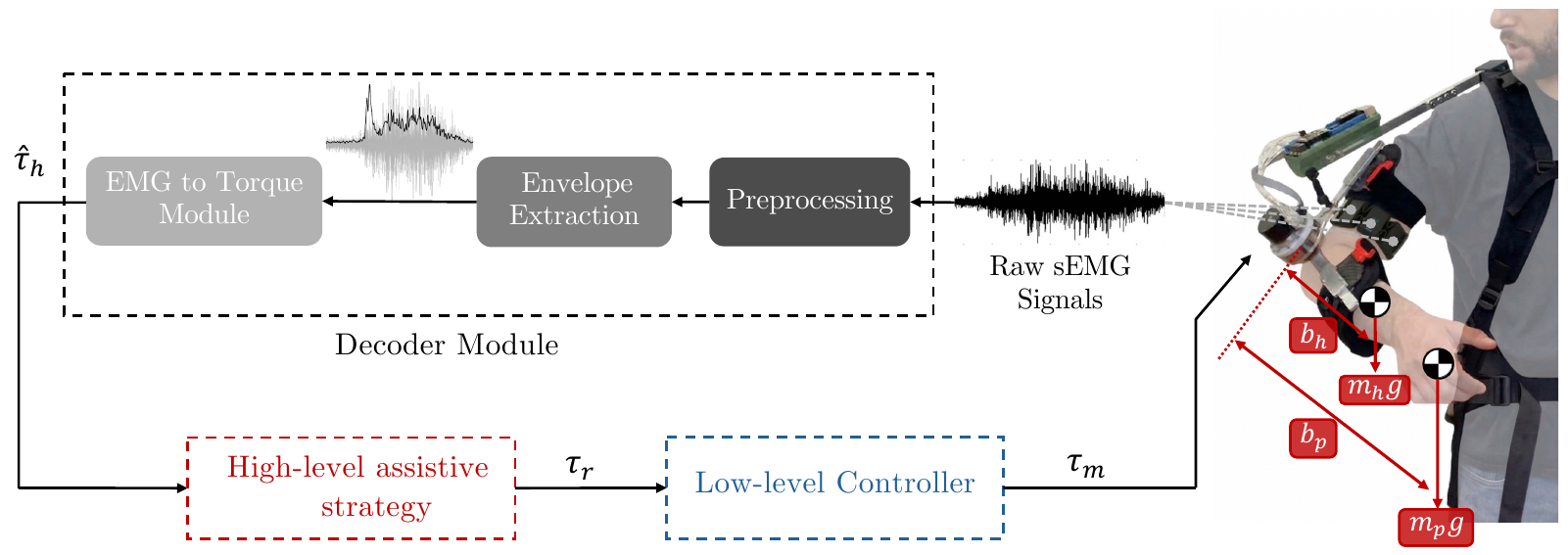}
    \caption{Architecture of an EMG-driven control system, implemented on a custom-built upper-limb exoskeleton prototype.}
    \label{fig:humanExo}
\end{figure}



EMG-driven exoskeleton control aims to translate the user’s neuromuscular activity into assistive torque delivered by the device. As shown in Fig. \ref{fig:humanExo}, the control pipeline starts with the acquisition of raw EMG signals from the upper arm. These signals are then processed through filtering, thresholding, and envelope extraction. The resulting activation signals are then typically provided to an EMG-to-torque module, which generates an estimate of the human joint torque $\hat{\tau}_h$ \cite{Quesada2026EMGToTorque}.

Between the estimated torque and the low-level control layer, different transformations can be introduced depending on the chosen assistive strategy. The estimated biological torque $\hat{\tau}_h$ can be amplified, reshaped, integrated, or used to adapt internal model parameters. In other words, the mapping between and the commanded assistive torque $\tau_m$ is algorithm-dependent and defines the high-level behavior of the exoskeleton. In this work, the comparative analysis focuses on this control block, and the specific control strategies considered are described in the following sections. Finally, the low-level controller is responsible for actuator control. In this paper, we focus on explicit force control, a widely adopted approach in wearable robotics, where the reference assistive torque $\tau_r$ is tracked to generate the actuator command $\tau_m$.

\section{EMG-driven Assistive Strategies}\label{sec:assistive_strategies}

\subsection{Proportional Myoelectric Control (PMC)}
\label{subsec:EMG:PMC}
Proportional myoelectric control (PMC) is the most established and widely adopted paradigm in EMG-driven exoskeletons, applied for both strength augmentation in healthy users \cite{rosen2001myosignal,lenzi2012:emg,missiroli2020:assistance_level,lotti2022:myo_vs_force,lotti2020:myocontrol_soft} and motor assistance in clinical populations \cite{Cheng2003:proportional_stroke,Song2008_proportional_stroke,costanzi2023:personal_myo}. 

\begin{figure}[h]
    \centering
    \includegraphics[width=0.95\columnwidth]{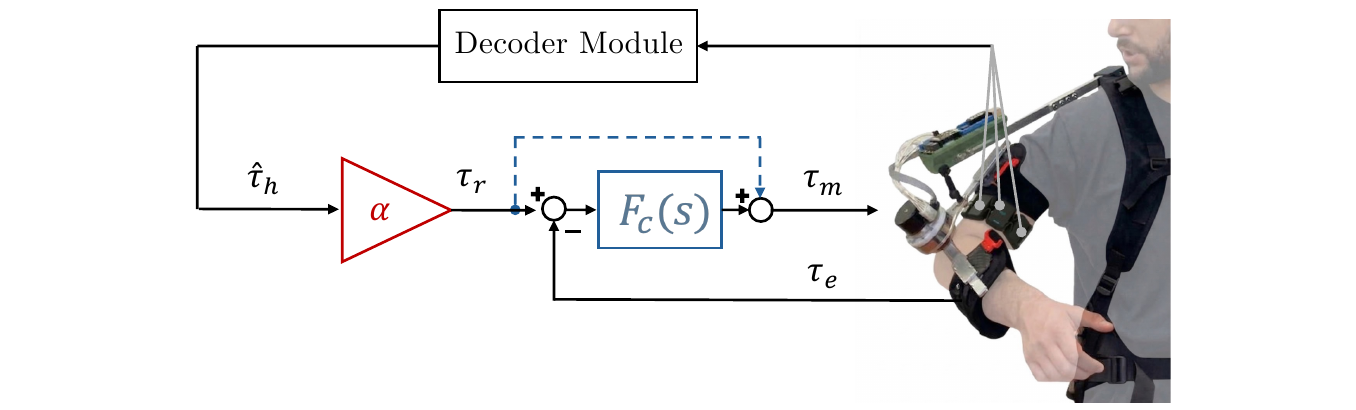}
    \caption{Block diagram of a Proportional Myoelectric Control (PMC) architecture.}
    \label{fig:proportional_control}
\end{figure}

As shown in Fig. \ref{fig:proportional_control}, the exoskeleton acts as a strength amplifier, providing an assistive torque proportional to the user's muscle activation through a force augmentation gain $\alpha$. Proportional control has become a benchmark in the field of myoelectric exoskeleton control \cite{lotti2022:myo_vs_force}. Notably, the only commercially available upper-limb exoskeleton that currently supports myoelectric control implements a proportional assistance strategy \cite{Peng2024Myomo}. Despite its popularity, proportional control presents intrinsic limitations. Assistance critically depends on gain tuning: insufficient amplification leads to limited support, while excessive gain can induce oscillatory behavior \cite{Kwon2014}.

\subsection{Adaptive Gravity Compensation: Pure Integral (AGC-PI)}
\label{subsec:EMG:AGCPI}

\begin{figure}[h]
    \centering
    \includegraphics[width=0.95\columnwidth]{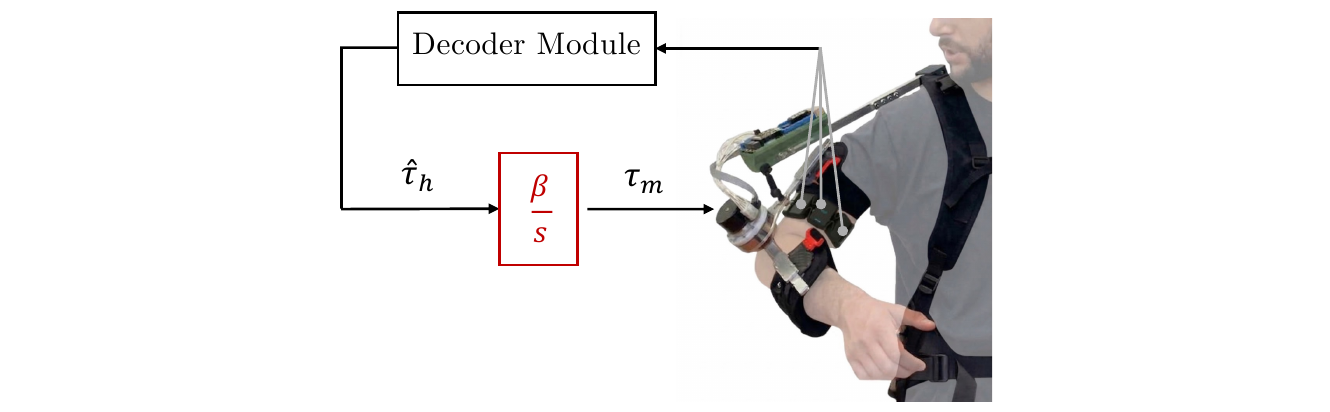}
    \caption{Block diagram of the Adaptive Gravity Compensation: Pure Integral (AGC-PI) architecture proposed in \cite{TreussartVignais2020}.}
    \label{fig:integral_control}
\end{figure}

One of the earliest implementations of adaptive gravity compensation based on a purely integral action was proposed in \cite{TreussartVignais2020}. Designed to assist lifting unknown loads without force sensing or prior payload knowledge, the control law integrates the estimated human torque to implicitly drive it toward zero in static conditions. A block diagram is shown in Figure \ref{fig:integral_control} where $\beta$ is the integral gain. As long as residual muscle activity counteracts gravity, the integrator increases the assistive torque, converging to a steady state once compensation is achieved. 

Because this formulation lacks an explicit gravitational model, the generated torque is independent of the joint angle, potentially compromising compensation across varying arm configurations. Furthermore, the absence of force sensing precludes the use of an explicit force control loop, impacting human-robot interaction transparency.

\subsection{Adaptive Gravity Compensation: Acceleration-Aware (AGC-AW)}
\label{subsec:EMG:AGCAW} 

\begin{figure}[t]
    \centering
    \includegraphics[width=0.95\columnwidth]{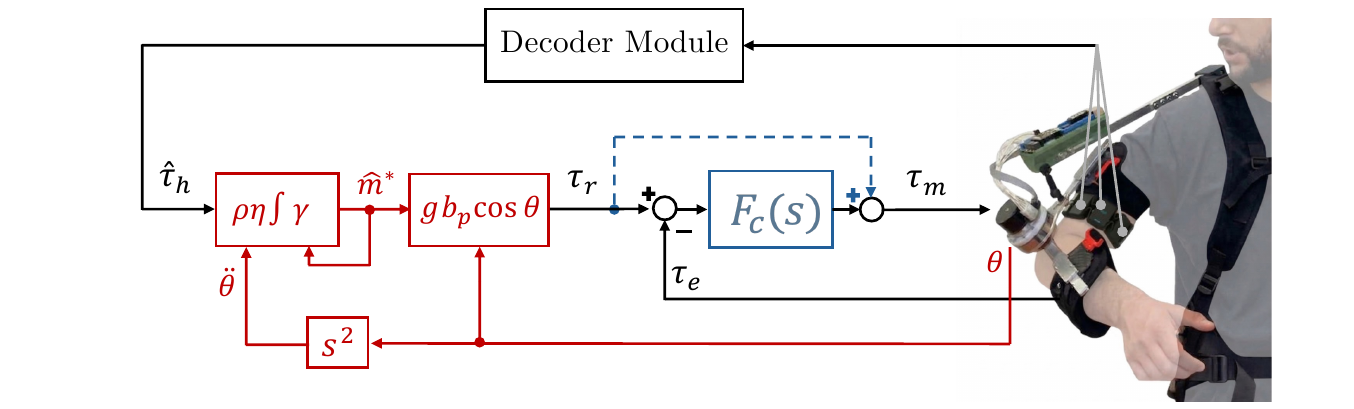}
    \caption{Block diagram of the Adaptive Gravity Compensation: Acceleration-Aware (AGC-AW) architecture proposed in \cite{Dimo2026MyographyGravityCompensation}.}
    \label{fig:acceleration_aware}
\end{figure}

A more recent, model-based evolution of the pure integral approach is the kinematics- and acceleration-aware AGC (\textit{AGC-AW}) proposed in \cite{Dimo2026MyographyGravityCompensation}, whose block diagram is represented in Fig. \ref{fig:acceleration_aware}. Unlike AGC-PI, this strategy explicitly incorporates the human-exoskeleton dynamic model. By projecting the inertial and gravitational dynamics onto a single equivalent mass $m^*$ located at the known payload distance $b_p$, the system dynamics in Eq. \eqref{eq:system} can be expressed as:
\begin{equation}\label{eq:eq_model}
    m^* b_p^2 \ddot{\theta} + m^* b_p g \cos\theta = \tau_h + \tau_e .
\end{equation}
Building upon this equivalent representation, AGC-AW mitigates the disturbances introduced by acceleration and anthropometric uncertainties during the mass estimation process. Furthermore, it integrates an explicit low-level force controller to enhance the transparency of the human-exoskeleton interaction.

The equivalent mass $m^*$ is estimated online through the following adaptive law:
\begin{equation}
    \hat{m}^* = \rho\,\eta\int\underbrace{\left(\hat{\tau}_h - \hat{m}^* b_p^2 \ddot{\theta}\right)}_{\gamma(t)}\,dt .
\end{equation}
Here, $\eta>0$ is the adaptation rate, and $\rho=\text{sign}(b_p\ddot{\theta}+g\cos\theta)$ accounts for configurations where muscle roles switch between pro-gravity and antigravity actions. To prevent acceleration-dependent torques from biasing the estimate, the inertial term $\hat{m}^*b_p^2\ddot{\theta}$ is actively subtracted from the estimated human torque within the integral.

Because this formulation can diverge near dynamic singularities or in kinematic configurations with negligible gravitational influence, the adaptation must be conditionally bounded \cite{Dimo2026MyographyGravityCompensation}. In its most robust variant (considered in this study), the mass update is suspended when joint acceleration is excessive or when the gravitational torque component is minimal:
\begin{equation}
\dot{\hat{m}}^*(t) =
\begin{cases}
    \rho\,\eta\, \gamma(t) & \text{if } |\ddot{\theta}| < \ddot{\theta}_{\text{th}} \text{ and } |\cos\theta| \geq \varepsilon_{\cos},\\[1mm]
    0 & \text{otherwise}.
\end{cases}
\end{equation}
where $\ddot{\theta}_{\text{th}} > 0$ is a predefined acceleration threshold, and $\varepsilon_{\cos} > 0$ is a small constant ensuring the update occurs only in configurations with significant gravitational excitation.

\subsection{Adaptive Gravity Compensation: Recursive Least Squares (AGC-RLS)}\label{subsec:EMG:AGCRLS}

\begin{figure}[h]
    \centering
    \includegraphics[width=0.95\columnwidth]{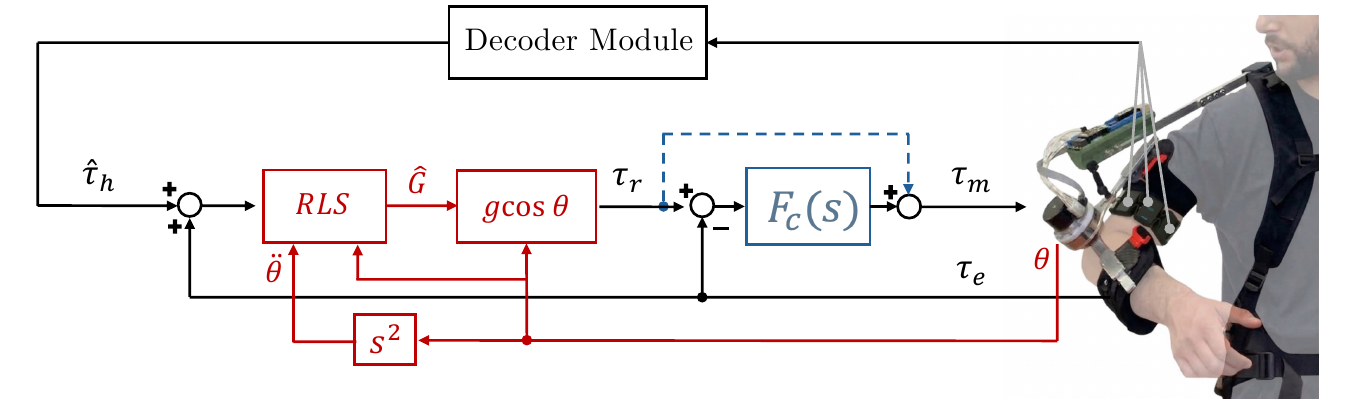}
    \caption{Block diagram of the Adaptive Gravity Compensation: Recursive Least Squares (AGC-RLS) architecture.}
    \label{fig:AGC-RLS}
\end{figure}

A contribution of this work is a novel MB AGC strategy based on Recursive Least Squares (RLS). Unlike previous AGC approaches, the payload is estimated through an online parametric identification of the elbow dynamics. A block diagram is shown in Fig. \ref{fig:AGC-RLS}.
System dynamics in Eq. \eqref{eq:system} can be represented as
\begin{equation}\label{eq:system_rls}
A\,\ddot{\theta} + G\, g \cos\theta = \tau_h + \tau_e,
\end{equation}
where $A = m_h b_h^2 + m_p b_p^2$ and $G = m_h b_h + m_p b_p$, have to be estimated. This model can be cast into a linear regression form
\begin{equation}
Y(k) = \varPhi^\top(k)\hat\Theta(k),
\end{equation}
with output $Y(k)=\hat{\tau}_h(k)+\tau_e(k)$, regressor $\varPhi(k) = [\ddot{\theta}(k), g\cos\theta(k)]^\top$ and parameter vector $\hat\Theta(k) = [\hat{A}(k), \hat{G}(k)]^\top$. The parameters are updated online using a standard RLS scheme
\begin{align}
\hat\Theta(k) &= \hat\Theta(k-1) + K(k)\,\varepsilon(k),\\
K(k) &= \frac{P(k-1)\varPhi(k)}{\lambda+\varPhi^\top(k)P(k-1)\varPhi(k)},\\
P(k) &= \frac{1}{\lambda}\left(P(k-1) - K(k)\varPhi^\top(k)P(k-1)\right),
\end{align}
where $\varepsilon(k) = Y(k) - \varPhi^\top(k)\hat\Theta(k-1)$ is the prediction error and $\lambda \in (0,1]$ is the forgetting factor ($K(k) \in \R^2$ and $P(k) \in \R^{2 \times 2}$).
The control reference $\tau_r$ is then set to
\begin{equation}\label{eq:tau_r_rls}
    \tau_r = \hat{G} g \cos\theta.
\end{equation}

It is worth noticing that system \eqref{eq:system_rls} can also be re-parametrized using system's COM distance $b'$ and its equivalent mass $m'$ as:
\begin{equation}\label{eq:system_prime}
    m' b'^2 \ddot{\theta} + m' g b' \cos{\theta} = \tau_h + \tau_e ,
\end{equation}
with $m'b'^2 = A$ and $m'b' = G$. The values of $\hat{m}'$ and $\hat{b}'$ can be obtained from $\hat{A}$ and $\hat{G}$ as
\begin{equation}\label{eq:params_rls}
    \hat{m}'=\frac{\hat{G}^2}{\hat{A}} \text{ and } \hat{b}' = \frac{\hat{A}}{\hat{G}} .
\end{equation}
Differently from equivalent mass $m^*$ of \textit{AGC-AW}, once the payload has been lifted, the parameters of model \eqref{eq:system_rls} are fixed, therefore $A$ and $G$ (or $m'$ and $b'$, considering \eqref{eq:system_prime}) are constant in time and do not present singularities, and this represents a key advantage.

\section{Experiments}\label{sec:experiments}

\subsection{Experimental Setup}\label{sec:exp_results:proc}

\begin{figure}[t]
    \centering
    \includegraphics[width=0.95\columnwidth]{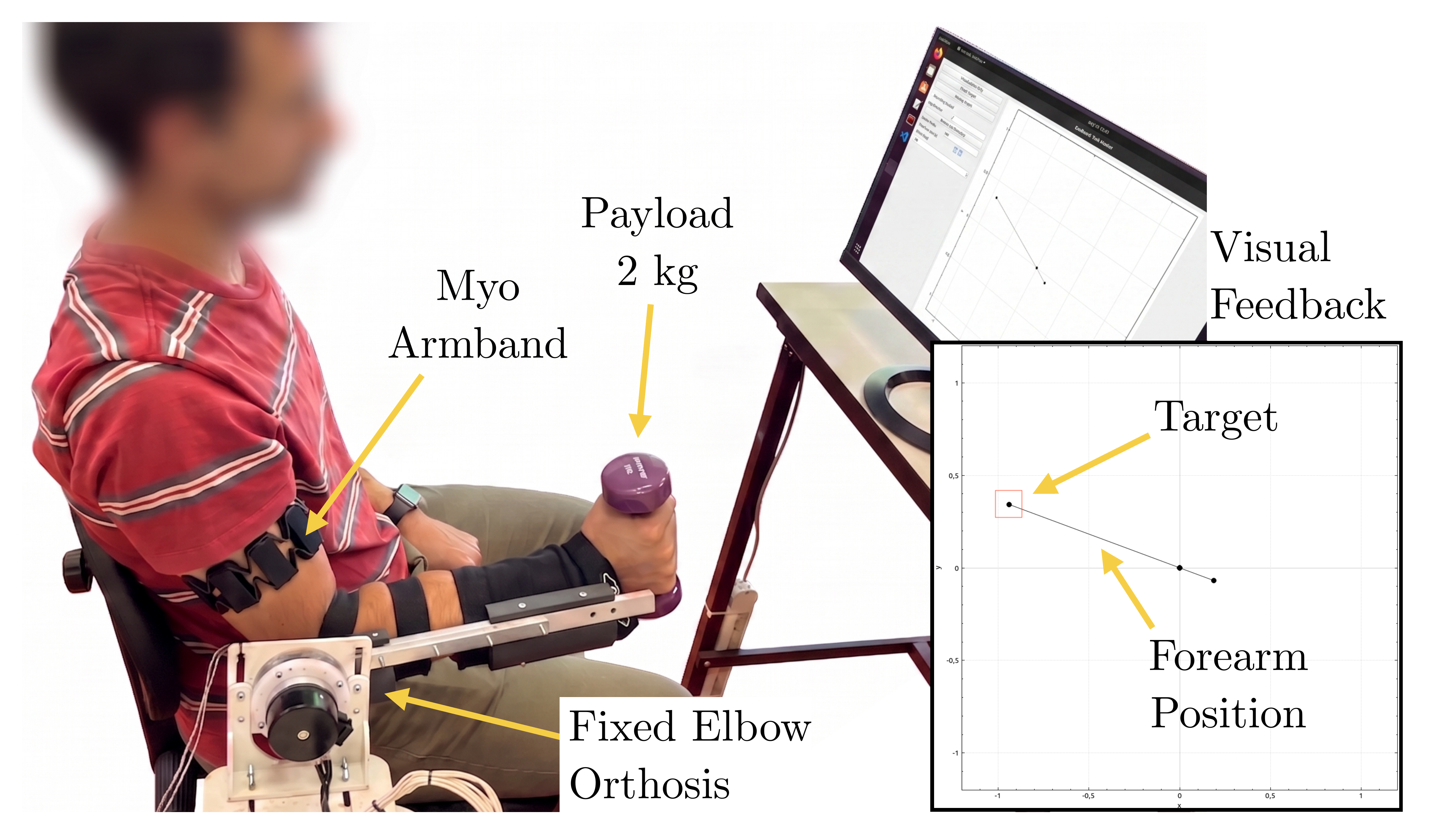}
    \caption{Experimental setup. Left: Participant performing the $2$~\unit{kg} loaded reaching task using a fixed 1-DoF robotic elbow orthosis and a Myo Armband for EMG acquisition. Right: Custom GUI providing real-time visual feedback of targets and forearm position.}
    \label{fig:exo-setup}
\end{figure}

The experimental setup (Fig. \ref{fig:exo-setup}) employs a fixed version of the custom-built 1-DoF elbow exoskeleton shown in Fig. \ref{fig:humanExo}, which was rigidly mounted to a table to isolate the control effects from wearability-related confounding factors and ungrounded reaction forces. Actuation is provided by a $100$~\unit{W} Maxon EC60 flat brushless motor coupled with a 1:81 reduction gear. Interaction with the user's forearm is secured via two adjustable strapped supports. The joint angle $\theta$ is measured via an optical encoder, with velocity estimated through hardware-timer measurements and acceleration computed via numerical differentiation and low-pass filtered at $5$~\unit{Hz}. Two strain gauges on the forearm directly measure the interaction torque $\tau_e$. Human torque estimation relies on EMG signals acquired from the upper arm via a Thalmic Labs Myo Armband. Signals are processed at $100$~\unit{Hz} in a ROS environment. Linear envelopes are extracted using a pipeline comprising a $50$~\unit{Hz} notch filter, a $20$~\unit{Hz} high-pass filter, full-wave rectification, and a $3$~\unit{Hz} second-order Butterworth low-pass filter.

\subsection{Assistive Control Strategies}
The control strategies described in Section~\ref{sec:assistive_strategies} were implemented and evaluated. The PMC strategy was tested at two force-augmentation levels: $\alpha = 0.5$ (\textit{PMC-50}) and $\alpha = 0.75$ (\textit{PMC-75}), the latter empirically identified as the maximum assistance level that allowed successful task completion across all participants. The three AGC strategies (\textit{AGC-PI}, \textit{AGC-AW}, and \textit{AGC-RLS}) were implemented as formally defined. With the exception of AGC-PI, the low-level explicit force controller $F_c(s)$ utilizes a PD structure with a feedforward term, tuned to ensure a $\sim 2$~\unit{Hz} closed-loop bandwidth. Key control parameters are summarized in Table \ref{tab:ctrl_par}.

\begin{table}[t]
\centering
\begin{tabularx}{\columnwidth}{
  c 
  X 
  c
}
  \toprule
  \bfseries Symbol & \bfseries Description & \bfseries Value \\
  \midrule
  $k_p$ & $F_c(s)$ proportional gain & $20.0$ \\
  $k_d$ & $F_c(s)$ derivative gain & $0.2$ \\[0.5ex]
  $\alpha$ & PMC force-augmentation level & $0.5$ and $0.75$ \\
  $\beta$ & \textit{AGC-PI} integral gain & $1.0$ \\
  $\eta$ & \textit{AGC-AW} adaptation rate & $1.0$ \\
  $\lambda$ & \textit{AGC-RLS} forgetting factor & $0.997$ \\
  \bottomrule
\end{tabularx}
\vspace{0.2cm} 
\caption{Control parameters used in the experimental trials.}
\label{tab:ctrl_par}
\end{table}

\subsection{Participants and Experimental Protocol}
Fourteen healthy, exoskeleton-naive subjects ($8$ females, $6$ males; age $20.3 \pm 0.5$ years; mass $70.4 \pm 12.3$~\unit{kg}; height $1.73 \pm 0.09$~\unit{m}) participated after providing written informed consent (University of Verona Ethics Committee, Protocol R1.21/2024).

The protocol initiated with an acclimation phase and anthropometric assessment, identifying the user-specific payload distance ($b_p = 0.35 \pm 0.03$~\unit{m}). To establish a baseline for EMG normalization, subjects performed three maximum voluntary isometric contractions (MVC) against static resistance for both elbow flexion and extension. Subsequently, the human joint torque $\hat{\tau}_h$ was calibrated. During a 1-minute static tracking task, the EMG-to-torque mapping was identified via linear least-squares regression, yielding an average fit of $R^2 = 0.86 \pm 0.04$.

For the dynamic task, participants interacted with a custom GUI (Fig. \ref{fig:exo-setup}) commanding 20 alternating flexion-extension reaching movements toward square targets ($4$-$6$~\unit{mm} side length). Target acquisition required a stable $1$~\unit{s} dwell time. Each algorithm was tested consecutively under an unloaded condition ($0$~\unit{kg}) and a loaded condition ($2$~\unit{kg} dumbbell). To mitigate order effects, the presentation sequence of the five control strategies was randomized, and the study was single-blinded. After completing both load conditions for a specific algorithm, participants evaluated their perceived workload using the NASA-TLX questionnaire.

\subsection{Evaluation Metrics}
Outcome was quantified per repetition $i$ using the following metrics:

\noindent  
\textbf{Muscular Activity:} The pre-processed EMG envelope was normalized to the MVC. Performance was evaluated as the mean normalized activation over the entire movement duration.

\noindent 
\textbf{Movement Accuracy:} Assessed via \textit{Overshoots} ($O_i$), a discrete count of instances where the end-effector entered the target area but exited before completing the $1$~\unit{s} dwell time.

\noindent 
\textbf{Task Efficiency:} Evaluated temporally via \textit{Completion Time} (total duration to acquire all 20 targets) and kinematically via \textit{Throughput} ($TP_i$). Based on Fitts' Law \cite{fitts1954}, $TP_i = ID_i / MT_i$, where $MT_i$ is the movement time and the Index of Difficulty is $ID_i = \log_2(1 + D_i/W_i)$ (in bits), with $D_i$ representing the angular distance and $W_i$ the target width.

\noindent 
\textbf{Subjective Workload:} Assessed via the six dimensions of the NASA-TLX \cite{hart1988nasatlx}.

\subsection{Statistical Analysis}\label{sec:stats}

All statistical analyses were performed using MATLAB 2025b, with significance set at $p < 0.05$. For continuous evaluation metrics (normalized Muscular Activity, Throughput, and Completion Time), Two-Way Repeated Measures (RM) ANOVAs evaluated the within-subject factors \textit{Algorithm} (5 levels) and \textit{Payload} (2 levels). Greenhouse-Geisser corrections were systematically applied to adjust degrees of freedom wherever Mauchly's test indicated sphericity violations.

Movement accuracy (Overshoots) was analyzed using a Generalized Linear Mixed-Effects Model (GLMM) with a Poisson distribution and a logarithmic link function. The model specified \textit{Algorithm}, \textit{Payload}, and their interaction as fixed effects, with random intercepts for \textit{Subject} and at the observation level to manage inter-subject variability and overdispersion. Non-significant interaction terms were removed prior to estimating marginal main effects.

Subjective workload (NASA-TLX ordinal scores) was assessed via the nonparametric Friedman test ($N = 14$), quantifying effect sizes with Kendall’s $W$. Across all aforementioned statistical models, significant main effects and interactions were subsequently investigated using Bonferroni-adjusted pairwise comparisons to control the Family-Wise Error Rate.

\section{Results}\label{sec:results}

\begin{table}[t]
\centering
\begin{tabularx}{\columnwidth}{
  l 
  X 
  c 
  c 
  c
}
\toprule
\bfseries Metric & \bfseries Effect & \bfseries F-Statistic & \bfseries df & \bfseries p-value \\
\midrule
\multirow{3}{*}{Biceps Brachii} 
& Algorithm & $31.16$ & $(4, 52)$ & $< 0.001$ \\
& Payload & $32.88$ & $(1, 13)$ & $< 0.001$ \\
& Alg. $\times$ Payl. & $24.82$ & $(4, 52)$ & $< 0.001$ \\
\midrule
\multirow{3}{*}{Triceps Brachii} 
& Algorithm & $11.18$ & $(4, 52)$ & $< 0.001$ \\
& Payload & $33.30$ & $(1, 13)$ & $< 0.001$ \\
& Alg. $\times$ Payl. & $0.70$ & $(4, 52)$ & $0.522$ \\
\midrule
\multirow{3}{*}{Throughput} 
& Algorithm & $42.35$ & $(4, 52)$ & $< 0.001$ \\
& Payload & $6.43$ & $(1, 13)$ & $0.025$ \\
& Alg. $\times$ Payl. & $3.10$ & $(4, 52)$ & $0.043$ \\
\midrule
\multirow{3}{*}{Completion Time} 
& Algorithm & $22.95$ & $(4, 52)$ & $< 0.001$ \\
& Payload & $1.81$ & $(1, 13)$ & $0.202$ \\
& Alg. $\times$ Payl. & $1.14$ & $(4, 52)$ & $0.338$ \\
\bottomrule
\end{tabularx}
\vspace{0.2cm} 
\caption{Two-Way RM ANOVA Results for Muscular Activity and Task Efficiency.}
\label{tab:anovares}
\end{table}

\begin{figure}[t]
    \centering
    \includegraphics[width=0.95\linewidth]{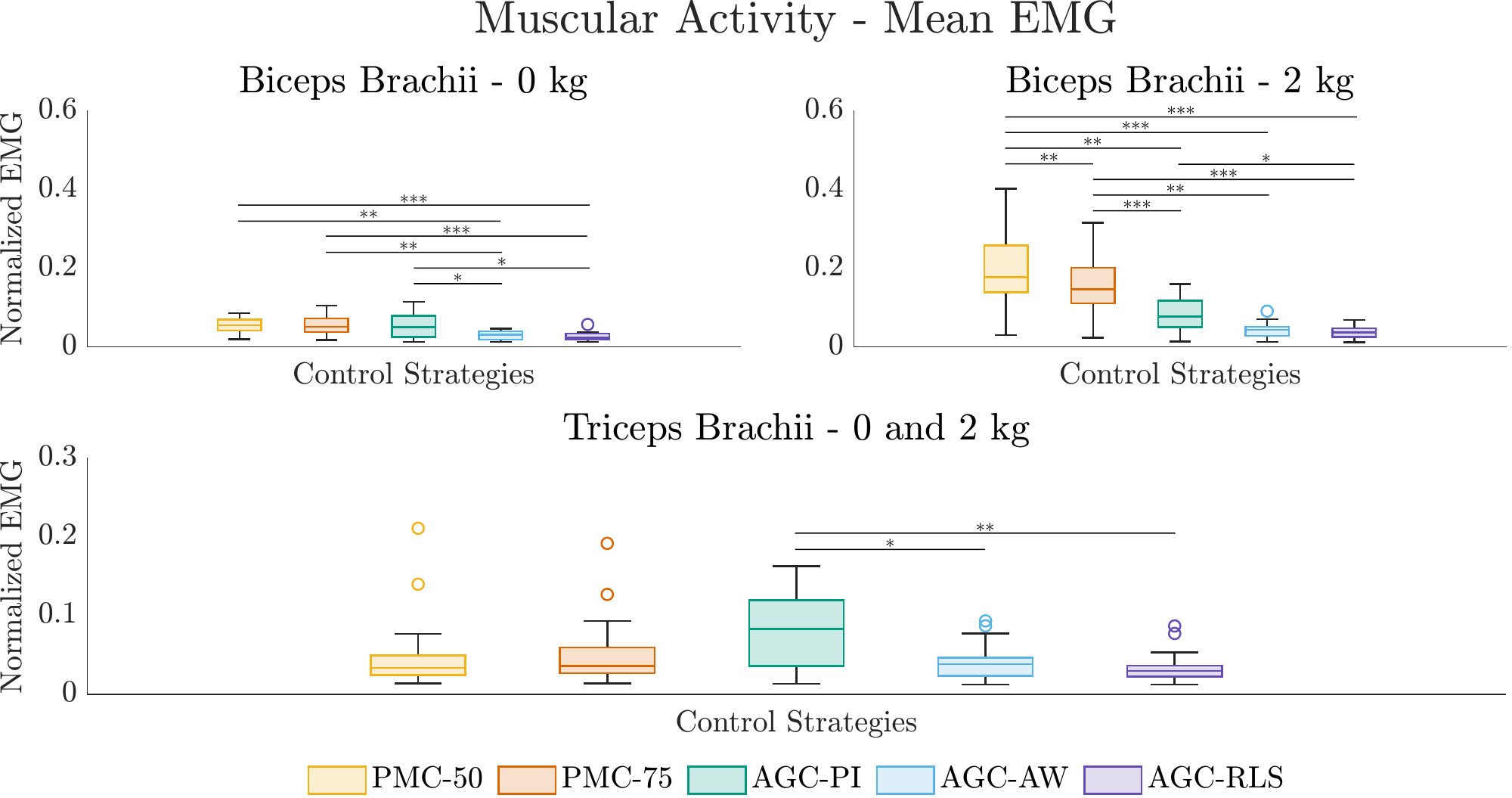}
    \caption{Muscular activity (Normalized Mean) for Biceps and Triceps Brachii. Biceps activation is shown for both 0~\unit{kg} and 2~\unit{kg} loads, while Triceps activation is aggregated.}
    \label{fig:emgres}
\end{figure}

\subsection{Muscular Activity}

Biceps Brachii activation (Table \ref{tab:anovares}, Fig. \ref{fig:emgres}) exhibited a significant \textit{Algorithm $\times$ Payload} interaction. In the unloaded condition ($0$~\unit{kg}), the kinematic-aware strategies (\textit{AGC-AW} and \textit{AGC-RLS}) elicited significantly lower activation than all other controllers ($p \le 0.048$), which remained statistically equivalent among themselves. Under the $2$~\unit{kg} load, this difference widened: \textit{AGC-AW} and \textit{AGC-RLS} maintained the lowest activation, significantly outperforming the PMC strategies (all $p \le 0.001$). Additionally, \textit{AGC-PI} resulted in significantly lower Biceps activation than both PMC variants ($p \le 0.001$).

Conversely, Triceps Brachii showed no payload interaction, with the algorithm determining the response entirely. \textit{AGC-PI} induced significantly higher Triceps activation compared to both \textit{AGC-RLS} ($p = 0.003$) and \textit{AGC-AW} ($p = 0.023$). The remaining algorithms demonstrated comparable, lower antagonist recruitment.

\subsection{Movement Accuracy}

\begin{figure}[t]
    \centering
    \includegraphics[width=0.95\linewidth]{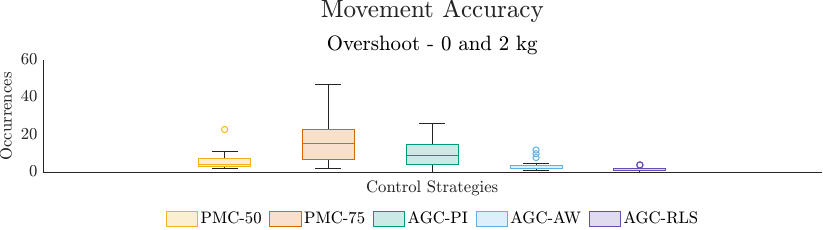}
    \caption{Number of overshoots observed for each control algorithm. Values are pooled across the 0~\unit{kg} and 2~\unit{kg} load conditions, as system stability was not significantly affected by the external weight.}
    \label{fig:accres}
\end{figure}

The GLMM evaluation indicated no significant \textit{Algorithm $\times$ Payload} interaction ($p > 0.17$); consequently, marginal main effects were estimated. The control algorithms demonstrated a strict hierarchy in dynamic stability (Table \ref{tab:posthoc_overshoot}, Fig. \ref{fig:accres}). \textit{PMC-75} recorded the highest overshoot frequency (all $p \le 0.027$ vs. other algorithms). Accuracy progressively improved with \textit{AGC-PI} and \textit{PMC-50} ($p = 0.010$ between them). The kinematic-aware strategies provided the highest stability, with \textit{AGC-RLS} significantly outperforming all other controllers, including \textit{AGC-AW} ($p = 0.002$).

\begin{table}[t]
\centering
\begin{tabularx}{\columnwidth}{
  l 
  c 
  c
  c
  c
  c
}
\toprule
 & \textit{PMC-50} & \textit{PMC-75} & \textit{AGC-PI} & \textit{AGC-AW} & \textit{AGC-RLS} \\
\midrule
\textit{PMC-50}  & -- & $< 0.001$ & $0.010$ & $0.021$ & $< 0.001$ \\
\textit{PMC-75}  & $< 0.001$ & -- & $0.026$ & $< 0.001$ & $< 0.001$ \\
\textit{AGC-PI}  & $0.010$ & $0.027$ & -- & $< 0.001$ & $< 0.001$ \\
\textit{AGC-AW}  & $0.021$ & $< 0.001$ & $< 0.001$ & -- & $0.002$ \\
\textit{AGC-RLS} & $< 0.001$ & $< 0.001$ & $< 0.001$ & $0.002$ & -- \\
\bottomrule
\end{tabularx}
\vspace{0.2cm} 
\caption{Cross-tabulation of Bonferroni-corrected $p$-values for pairwise comparisons of Overshoot occurrences between control strategies.}
\label{tab:posthoc_overshoot}
\end{table}

\subsection{Task Efficiency}

\begin{figure}[b]
    \centering
    \includegraphics[width=0.95\linewidth]{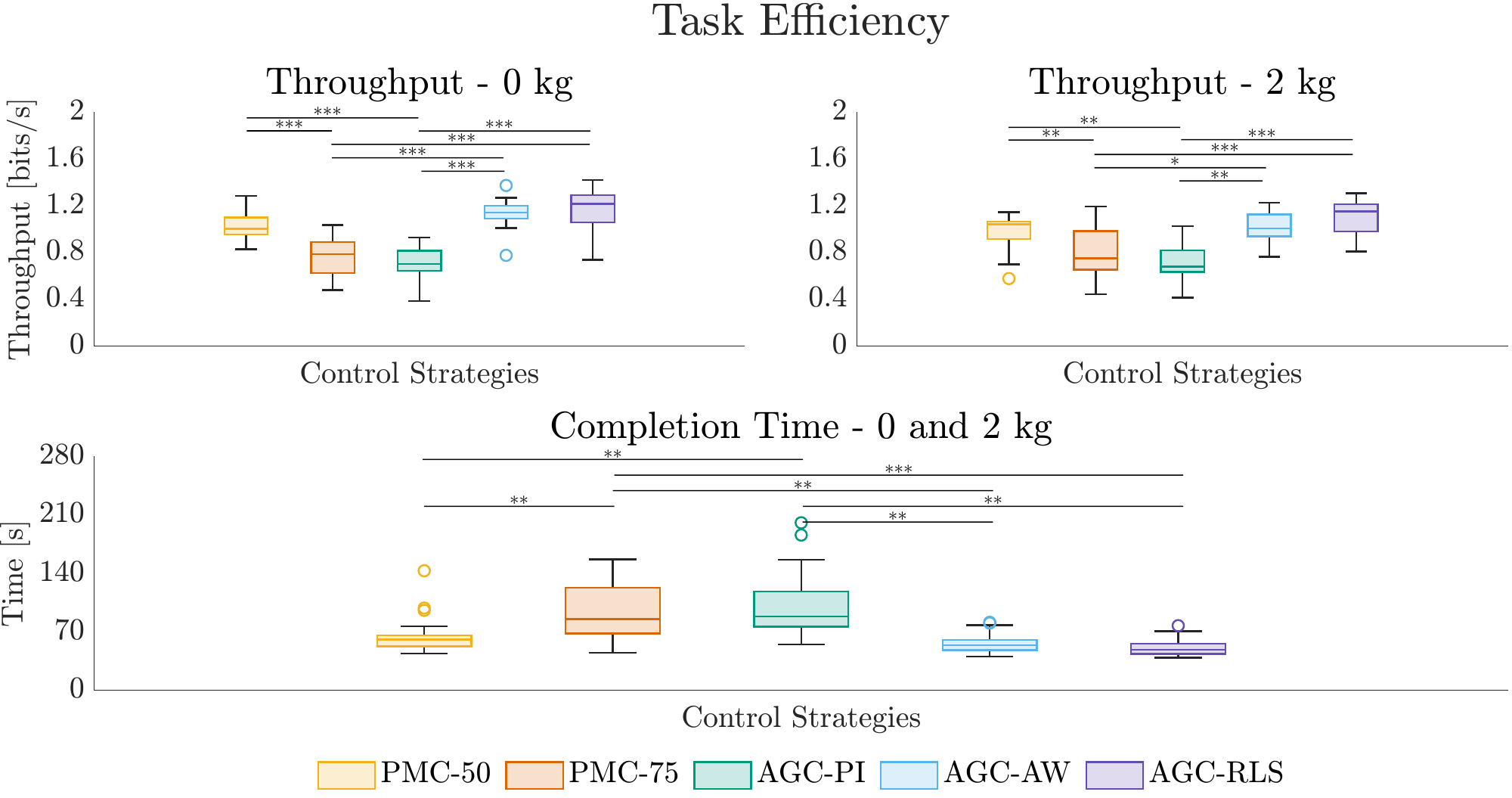}
    \caption{Throughput and Completion Time for each control algorithm. Throughput is shown for both 0~\unit{kg} and 2~\unit{kg} loads, whereas Completion Time is pooled across conditions.}
    \label{fig:perres}
\end{figure}

Throughput (Table \ref{tab:anovares}, Fig. \ref{fig:perres}) exhibited a significant \textit{Algorithm $\times$ Payload} interaction ($p = 0.043$). \textit{AGC-AW}, \textit{AGC-RLS}, and \textit{PMC-50} yielded statistically equivalent throughput across all tests. In the unloaded condition, these three algorithms outperformed \textit{PMC-75} and \textit{AGC-PI} (all $p < 0.001$). Under the $2$~\unit{kg} load, although performance gaps narrowed, \textit{AGC-AW}, \textit{AGC-RLS}, and \textit{PMC-50} maintained significantly higher throughput than \textit{PMC-75} and \textit{AGC-PI} (all $p \le 0.020$).

Completion Time was robust to external weight, showing no significant main effect of payload or interaction (all $p \ge 0.202$). Collapsing across load conditions, \textit{AGC-AW}, \textit{AGC-RLS}, and \textit{PMC-50} achieved statistically equivalent and significantly faster completion times compared to \textit{PMC-75} and \textit{AGC-PI} (all inter-group $p \le 0.007$).

\subsection{Cognitive and Physical Demand}

\begin{table}[t]
\centering
\begin{tabularx}{\columnwidth}{
  X 
  c 
  c 
  c
}
\toprule
\bfseries NASA-TLX Dimension & \bfseries $\chi^2$ & \bfseries p-value & \bfseries Kendall's $W$ \\
\midrule
Mental Demand & $28.24$ & $< 0.001$ & $0.504$ \\
Physical Demand & $18.93$ & $< 0.001$ & $0.338$ \\
Temporal Demand & $15.82$ & $0.003$ & $0.282$ \\
Performance & $19.79$ & $< 0.001$ & $0.353$ \\
Effort & $18.82$ & $< 0.001$ & $0.336$ \\
Frustration & $20.43$ & $< 0.001$ & $0.365$ \\
\bottomrule
\end{tabularx}
\vspace{0.2cm}
\caption{Friedman Test Results for NASA-TLX Subscales ($N=14$, $df=4$). Effect size is reported as Kendall's $W$.}
\label{tab:tlxres}
\end{table}

\begin{figure}[t]
    \centering
    \includegraphics[width=0.95\linewidth]{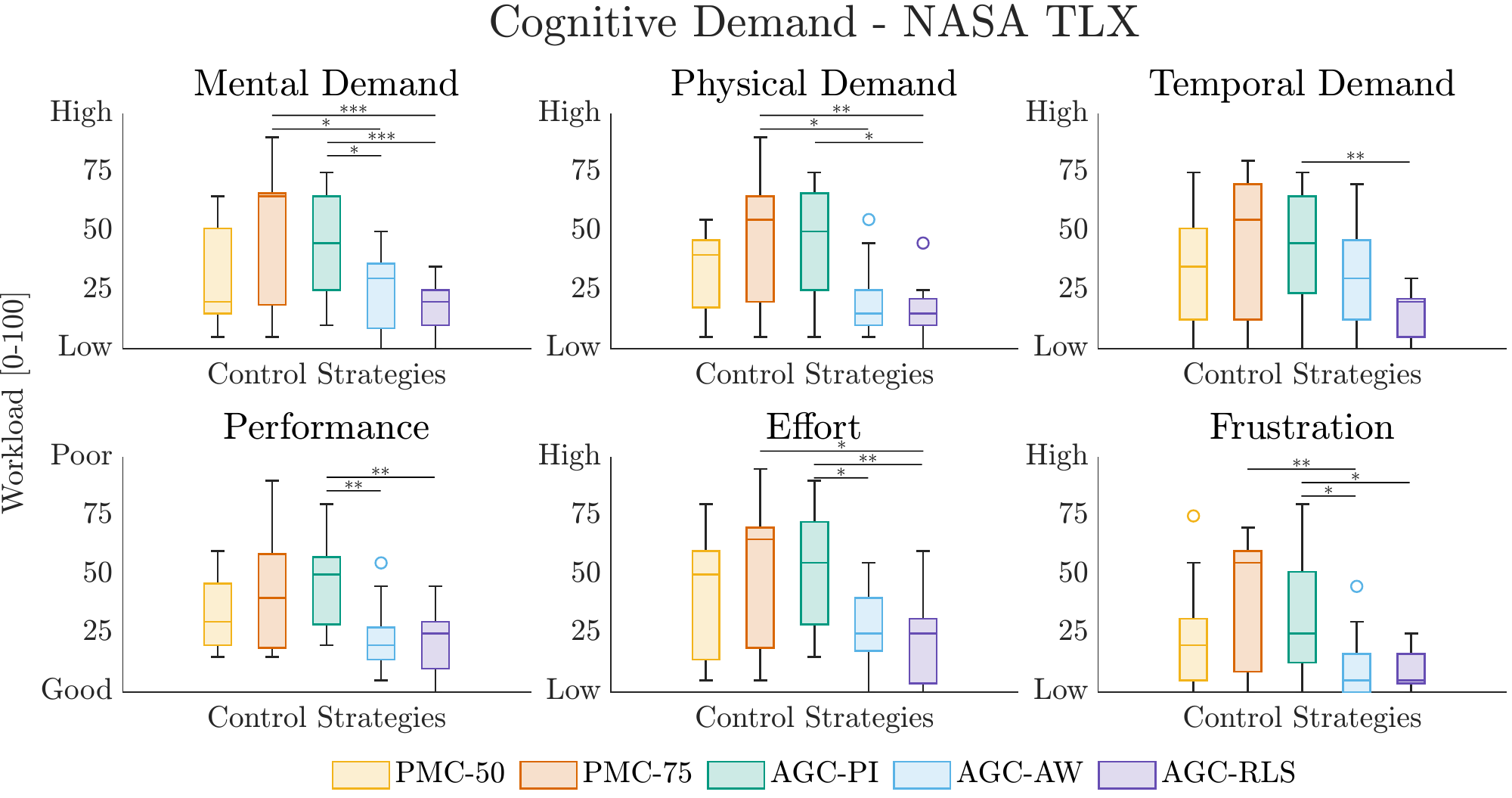}
    \caption{NASA-TLX subjective workload scores across the six subscales. Lower scores indicate lower perceived demand, except for Performance, where lower scores indicate better perceived performance.}
    \label{fig:tlxres}
\end{figure}

NASA-TLX scores (Table \ref{tab:tlxres}, Fig. \ref{fig:tlxres}) revealed a significant main effect of the control algorithm across all six subscales (all $p \le 0.003$, Kendall's $W \ge 0.282$).

\textit{PMC-50} exhibited no statistically significant differences in perceived workload compared to \textit{AGC-AW} and \textit{AGC-RLS} across any dimension (all $p \ge 0.127$). In contrast, \textit{AGC-PI} was rated significantly worse than both \textit{AGC-AW} and \textit{AGC-RLS} in Mental Demand, Performance, Effort, and Frustration (all $p \le 0.048$), and imposed significantly higher Physical and Temporal demands than \textit{AGC-RLS} (all $p \le 0.034$). Similarly, \textit{PMC-75} elevated Physical and Mental Demand compared to the MB algorithms (all $p \le 0.014$), required greater Effort than \textit{AGC-RLS} ($p = 0.012$), and induced higher Frustration than \textit{AGC-AW} ($p = 0.006$).

\section{Discussion}\label{sec:discussion}

A primary finding of this study is that kinematic precision, temporal efficiency, and user workload are determined primarily by the control algorithm rather than the handled payload. A comprehensive assessment of the evaluated strategies highlights a pronounced discrepancy between objective task execution and the associated biomechanical and cognitive costs.

Rather than lacking adaptability, PMC presents a structural limitation: it imposes a strict assistance-level trade-off between physical workload reduction and effectiveness. Regarding temporal efficiency, \textit{PMC-50} clustered with the advanced AGC strategies, achieving the fastest completion times. However, this speed came at a high physical cost: under the $2$~\unit{kg} loaded condition, \textit{PMC-50} elicited a disproportionate increase in Biceps Brachii activation. This indicates that users were forced to actively compensate for the controller's insufficient assistance to maintain their execution speed. Conversely, attempting to reduce this physical effort by increasing the assistance gain (\textit{PMC-75}) significantly worsened task execution. The marginal stability observed with \textit{PMC-75}, characterized by longer completion times and a significantly higher number of target overshoots, directly translated into the elevated levels of \textit{Frustration} and \textit{Mental Demand} reported in the NASA-TLX.

The \textit{AGC-PI} strategy addressed only part of this limitation. This strategy reduced Biceps Brachii activation compared with PMC under load, but it also led to significant antagonist co-contraction, as evidenced by consistently higher Triceps Brachii activation. From a motor control perspective, increasing joint stiffness via co-contraction is a typical human strategy to stabilize unpredictable or suboptimal external dynamics. This compensatory mechanism significantly increases the metabolic cost and discomfort, aligning with the lower self-assessed \textit{Performance} and higher \textit{Effort} reported by users. Importantly, \textit{AGC-PI} was the only strategy tested without a low-level explicit force controller. As a result, actuator friction and inertia could not be effectively masked at the interaction level and were therefore more directly perceived by the users. This may have contributed to the higher NASA-TLX scores observed for \textit{AGC-PI}.

In contrast, the model-based AGC strategies successfully broke the trade-off inherent to proportional control, providing the best overall balance across the evaluated metrics. Both \textit{AGC-AW} and \textit{AGC-RLS} provided maximal assistance while maintaining low muscular activation, achieving high throughput, short completion times, and a low number of overshoots. Their performance was also robust across payload conditions: the introduction of the $2$~\unit{kg} load did not significantly degrade kinematic accuracy or increase muscle activation. This robust performance is directly linked to their capacity to accurately identify the system dynamics online. During the loaded trials, both adaptation laws successfully converged near the nominal external payload, with \textit{AGC-RLS} estimating an added mass of $1.93 \pm 0.23$~\unit{kg} and \textit{AGC-AW} estimating $1.90 \pm 0.15$~\unit{kg}. This confirms accuracy of both adaptation algorithms.

\subsection{Limitations and Future Work}
While the use of a 1-DoF exoskeleton rigidly mounted to a table successfully isolated the fundamental effects of the control strategies, a primary limitation of this study lies in the lack of component-level isolation within the algorithms. The controllers were implemented according to their original literature formulations. Consequently, it is not possible to definitively attribute the failure of the \textit{AGC-PI} strategy or the success of the MB approaches to the use of distinct technologies, such as the presence of an explicit force control loop, or to the integration of model-based mechanisms.

Furthermore, the current experimental protocol and discrete reaching tasks did not expose significant performance divergences between the \textit{AGC-AW} and \textit{AGC-RLS} strategies. However, distinct limitations exist between the two formulations. The efficacy of \textit{AGC-AW} relies heavily on the empirical tuning of heuristic thresholds ($\varepsilon_{\cos}$, $\ddot{\theta}_{\text{th}}$), the adaptation rate ($\eta$), and an accurate a priori measurement of the payload distance $b_p$. While \textit{AGC-RLS} was designed to overcome these dependencies by estimating a time-invariant parametric model, the standard reaching task failed to highlight this structural advantage. The superiority of the RLS formulation is expected to emerge primarily in tasks involving continuous, dynamic payload variations during movement (e.g., pouring liquids or handling shifting loads), rather than a constant-payload task tested here.

Therefore, future research must design highly dynamic tasks with continuous load variations to expose the true operational limits of these estimators. Additionally, it will be necessary to validate these paradigms in ungrounded multi-DoF systems, and test them within clinical populations to fully establish their translational potential.

\section{Conclusion}\label{sec:conclusion}

This study provided a systematic, human-in-the-loop comparison of the proportional myoelectric control (PMC) and adaptive gravity compensation (AGC) paradigms for EMG-driven exoskeletons. While PMC remains the most widely adopted approach, our findings highlight its fundamental limitations under varying external loads, revealing a strict trade-off between physical effort reduction and dynamic stability. Similarly, purely integral AGC strategies proved biomechanically inefficient due to excessive co-contraction of antagonists.

Conversely, incorporating kinematic and acceleration awareness into the adaptation law—as demonstrated by the \textit{AGC-AW} and our novel \textit{AGC-RLS} strategy—effectively resolves these limitations. By performing fast, online payload estimation, these controllers successfully decoupled task efficiency from physical and cognitive fatigue, maintaining robust kinematic precision, temporal efficiency, and minimal muscle activation regardless of the handled weight.

\section*{Acknowledgment}
The authors would like to acknowledge the support of PR Veneto FESR 2021--2027 through the SHIELD project, ``Sustainable, Human-centric, Innovative, Efficient Logistics through Digital Advancements'' -- Project 24729\_002199.

\bibliographystyle{IEEEtran}
\bibliography{IEEEabrv,bibreport}


\end{document}